# Residual Bi-Fusion Feature Pyramid Network for Accurate Single-shot Object Detection

[1]*Ping-Yang Chen*, [2]*Jun-Wei Hsieh*, [3]*Chien-Yao Wang*, [3]*Hong-Yuan Mark Liao, and* [4]*Munkhjargal Gochoo*
[1]Department of Computer Science, National Chiao Tung University, Taiwan
[2]College of Artificial Intelligence and Green Energy, National Chiao Tung University
[3]Institute of Information Science, Academia Sinica
[4]Department of Computer Science & Software Engineering, United Arab Emirates University

pingyang.cs08g@nctu.edu.tw, jwhsieh@nctu.edu.tw,
x102432003@yahoo.com.tw, liao@iis.sinica.edu.tw, mgochoo@uaeu.ac.ae

## Abstract

*State-of-the-art (SoTA) models have improved the accuracy of object detection with a large margin via a FP (feature pyramid). FP is a top-down aggregation to collect semantically strong features to improve scale invariance in both two-stage and one-stage detectors. However, this top-down pathway cannot preserve accurate object positions due to the shift-effect of pooling. Thus, the advantage of FP to improve detection accuracy will disappear when more layers are used. The original FP lacks a bottom-up pathway to offset the lost information from lower-layer feature maps. It performs well in large-sized object detection but poor in small-sized object detection. A new structure "residual feature pyramid" is proposed in this paper. It is bidirectional to fuse both deep and shallow features towards more effective and robust detection for both small-sized and large-sized objects. Due to the "residual" nature, it can be easily trained and integrated to different backbones (even deeper or lighter) than other bi-directional methods. One important property of this residual FP is: accuracy improvement is still found even if more layers are adopted. Extensive experiments on VOC and MS COCO datasets showed the proposed method achieved the SoTA results for highly-accurate and efficient object detection..*

1. Introduction

Recently, the accuracy of object detection models have been improved by a large margin with various state-of-the-art (SoTA) models like FPN [1], YOLOv3 [2], and SSD [3], which usually consist of deep feature extractors (with backbones such as DarkNet-53 [4]) and ResNet-101[5], a feature pyramid (FP), and a classifier. Using a small or medium-sized backbone can reduce the computational cost and thus increase efficiency. However, a shallow backbone usually means a reduced number of rich semantic features for object detection. To increase the accuracy, most of SoTA methods adopt a very deep CNN structure to detect objects which leads a smaller object (<32× 32 pixels) to become a one-pixel feature in the last layer of the feature pyramid, one-pixel is insufficient for accurate object discrimination. For each backbone, followed by a convolution layer is a pooling method for reducing spatial resolution of a feature map. However, the pooling technique is not shift-invariant [6]. That is, shifts in an input will change the pooling output and also degrade the accuracy on small object detection. As describe in [7], the above state-of-the-art methods show impressive results on large and medium sized objects, but perform quite poorly on small objects. Small object detection is challenging and requires both high-level semantics to discriminate objects from the background and low-/mid-level features for accurate object localization. For YOLO v3, in order to improve the accuracy of small object detection, detailed features from more grids should be maintained and also degrade the results on large sized object detection. Thus, the winner team in LPIRC 2019 [28] improved the overall accuracy of the COCO dataset by enhancing the results on sizeable sized and medium-sized objects but ignoring the ones on small objects.

A feature pyramid (FP) structure is commonly adopted in state-of-the-art detectors for detecting objects at different scales. It extracts spatial features from the last feature layer so that semantically strong features can be maintained along a top-down path and lead to significant accuracy improvements on object detection. This top-down aggregation is now a commonly adopted design choice to improve scale invariance in both two-stage and one-stage detectors. There are few common types of FPs employed in object detection models, *i.e.*, pyramidal feature hierarchy (bottom-up), hourglass (bottom-up and top-down), FPN [1], SPP [8], and PFPN[9], respectively. However, the top-down pathway cannot preserve their accurate positions due to the shift-effect of pooling. To convey accurate localization, a bottom-up pathway is needed to offset the lost information from lower-layer feature maps. In [10], Woo et al. proposed a gated bidirectional feature pyramid network to tackle this issue by using a gating module on the SSD frame. The gate module is not easy to be trained. In [7], Wang et al. proposed a bi-directional network to efficiently circulate both low-/mid-level and high-level semantic information in the detection framework. However, an extra light-weight scratch network (LSN) should be trained to get a down-sampled image as input to efficiently construct low-/mid-level features.

To maintain both results on large and small-sized object detection, a new structure ReBiF (Residual BiFusion) feature pyramid is created in this paper. It is bidirectional and can fuse both deep and shallow features towards more effective and robust object detection. Due to the "residual" nature, similar to ResNet [5], it can be easily trained and integrated into different backbones (even deeper or lighter) than other bi-directional methods [7], [10]. Besides this structure, a new BiFusion module is proposed to let the "residual" features form a compact representation that brings more accurate localization information into each prediction layer so that not only the results on small-sized object detection but also large/medium-sized ones are improved. The RBF feature pyramid is simple, efficient, and can be easily applied to mobile (or embedded) applications since only features from adjacent three layers are fused to construct each prediction map with simple concatenations and 1×1 convolutions. Better generalizations of the proposed modules are proved by evaluating their performances across different backbones, object classes, and object sizes for general object detection on the VOC and MS COCO datasets. As a summary, our main contributions include the following:

- A novel residual bi-fusion feature pyramid is proposed to fuse both features from deep and shallow layers toward more accurate one-shot object detection.
- The new feature pyramid can be easily trained and integrated into different backbone due to its "residual" nature.
- Better generalization of the residual feature pyramid is achieved when evaluating its performance on different data sets, object sizes, and object classes.
- The BiFusion module can reduce the shift-invariance effect of pooling on object detection.
- Extensive experiments on VOC and MS COCO datasets [43] showed the proposed method achieved the SoTA results for highly-accurate and efficient object detection.

2. Related Works

Currently, object detection models in the literature can be classified into two approaches; a two-stage (proposal driven) and a one-stage (direct) approach. Generally, the former is known for its high accuracy and the latter is known for its efficiency. To investigate an efficient single-shot detection scheme towards high accuracy, this work focuses on one stage object detector.

2.1. One-stage object detectors

One-stage object detector consists of a backbone network (referred to backbone) and a predictor. The backbone is a stacked feature map that represents an input image in a high-feature resolution or in a low-spatial resolution. Mostly, the backbone is pre-trained as a single image classifier on a very large dataset, i.e., ImageNet. In 2013, the first CNN-based one-stage object detector OverFeat [11] was developed using a sliding-window paradigm. Then, two years later, YOLO [12] achieved the state-of-the-art performance by integrating bounding box proposal and subsequent feature resampling in one stage. Next, SSD ([3] employed in-network multiple feature maps for detecting objects with varying shapes and sizes, and the maps make SSD more robust than YOLOv1. For better detection of small objects, FPN [1] is developed using a feature pyramid (FP) structure and it achieves a higher detection accuracy on small objects. Later, the state-of-the-art YOLOv3 [2] was developed by adopting the concept of FPN. By changing the backbone from DarkNet-19[4] to DarkNet-53 [2], YOLO v3 achieves the best performance. Similarly, RetinaNet [13], a combination of FPN and ResNet as a backbone, proposes the use of focal loss to significantly reduce false positives in one-stage detectors by dynamically adjusting the weights of each anchor box.

Recently, there have been some frameworks proposed to evaluate the effects of small geometry perturbations on CNNs. In CNNs, shift-invariance is achieved with subsampling layers. However, later works [6] have found the max-pooling is more effective in object detection and classification. Thus, the subsampling operation in CNNs has been replaced with a max pooling operation. In [14], Zhang proposed a Pooling-after-Blurred technique by combing blurring and subsampling techniques to assure shift-invariance.

2.2. Latest one-stage object detectors

Object detection is one of the most popular fields in computer vision, and several state-of-the-art models have been developed in the past few years. RefineDet [15] employed an Encode-Decode structure for deepening the network and up-sampling deeper scale features to the shallower scales to enrich the contextual information for the final FP. A newly proposed PeLee model [16], a variant of DenseNet [17], outperformed SSD+MobileNet by 6.53% on Stanford Dogs dataset [18] with its much shallower network. However, PeLee has a lower performance on MS COCO [43] dataset and lower accuracy on small object detection. PFPN [9] adopts VGGNet-16 [19] as a backbone and SPP for generating a final FP from the last layer of a backbone which concatenates multi-scale features. The above mentioned model outperformed other methods on small object detection. To avoid the problem of handcrafted anchors, some anchor-free methods [20]-[23] were proposed. For example, in [20], corner features are first detected to anchor information for precise object detection. In [21], a cascade anchor refinement module is proposed to gradually refine predesigned anchors and then injected to a bidirectional pyramid network to detect objects with highly accurate locations. During training, only regression once is not good enough for accurate object detection for a one-stage object detector. In [24], a hierarchical shot detector is proposed to output the true boxes of detected objects via regression. However, the regression method is more accurate but inferiorly efficient.

To offset the lost information from shallow layers, there are more methods [7], [10], [21] adopting a bidirectional feature pyramid to enhance the accuracy of small object detection. In [10], a gating module was proposed to control the direction of feature flows to tackle this issue. In [7], Wang *et al*. constructed a light-weight scratch network and a bi-directional network to efficiently circulate both low-/mid-level and high-level semantic information for better small object detection. The latest state-of-the-art one-stage object detector M2Det [25] outperformed all the existing methods on all multi-scale categories on MS COCO. However, its model is complicated, time-consuming, and not suitable for a real-time object detection task due to its high computational cost.

## 3. Methods

FPN is a top-down method to bring semantically strong features from the last layer to discriminate objects from the background (see Figure 1(a)). However, it cannot preserve their accurate positions due to the effect of pooling and quantization. To tackle this problem, it is better to predict objects not only from the current layer but also shallow layers to prohibit small objects from being lost. Most of SoTA bi-directional methods create new feature streams from raw images or low-level feature layers to achieve accurate object localization. This scheme is memory and bandwidth-consuming for most embedded devices. As shown in Figure 1(b), we propose a **CORE** (**CO**ncatenation and **RE**organization) module to circulate semantic and localization information by reusing feature maps only from three layers (previous, current, and next) of the backbone. The "re-using" mechanism makes it be memory- and-bandwidth saved, and suitable for embedded applications. The output of the CORE module can be further "purified" to generate more semantic features. When the "purification" module is injected into the feature pyramid, a new Bi-Fusion feature pyramid is constructed for high-quality object detection (see Figure 1(c)). Inspired by the concept of ResNet-101[5], this pyramid can be easily, efficiently, and effectively trained if the "residual" concept is implemented. Figure 1(d) shows the final architecture of this residual feature pyramid.

Details of each component are described as follows.

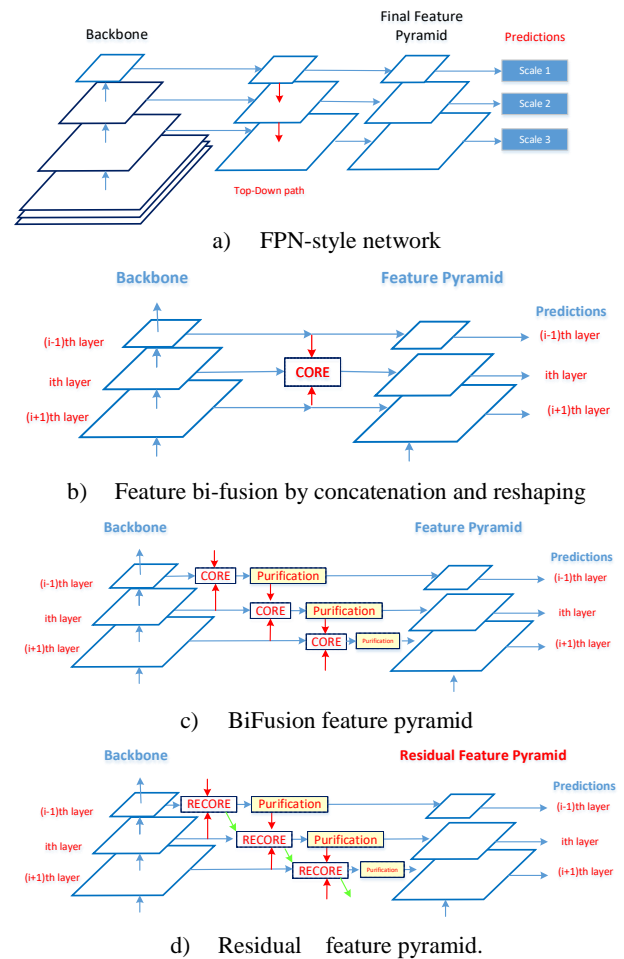

a) FPN-style network

b) Feature bi-fusion by concatenation and reshaping

c) BiFusion feature pyramid

d) Residual feature pyramid.

Figure 1. Architectures of a) SPP based FP network, b) feature bifusion by concatenation and reshaping, c) Bi-fusion feature pyramid, and (d) Residual feature pyramid.

### 3.1. CORE Module for Feature Fusion

One novelty of this work is a **CORE** module which can be recursively executed to not only concatenate high-level semantic features from deep layers to shallower layers (top-down direction) but also re-organize spatially richer features from a shallower layer to a deeper layer (bottom-up direction). To avoid using too many dithering operations (*i.e.*, convolutions) and computationally expensive operations (*i.e.*, pooling and addition) to preserve as much as possible features for prediction, the **CORE** module (see Figure 2) adopts concatenation for fusing features of a deeper layer to a current layer and reorganization operation for fusing features of a shallower layer to a current layer. Unlike concatenation methods used in state-of-the-art methods, the proposed CORE block recursively concatenates contextual features of not only adjacent layers but also a deeper layer. In other words, the CORE block fuses various features from 4 adjacent scales (shallow, current, deep, and more in-depth) of a backbone to richen the features for better detection. Both operations are time-efficient and can preserve all contextual information. Under these circumstances, accuracy and efficiency are both improved.

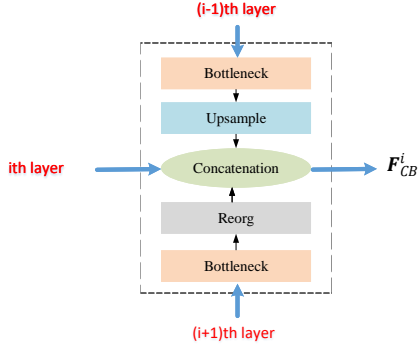

Figure 2. The CORE module for recursively concatenating contextual features from adjacent layers (previous, current, net).

### 3.2. Purification

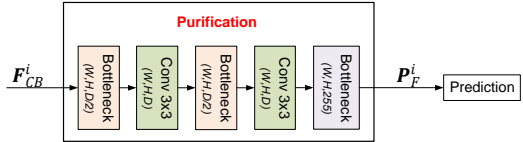

Figure 3. The "Purification" module.

The output of the CORE module can be further purified to form more contextual and semantic features from the fused features of 4 adjacent scales. Figure 3 illustrates the flowchart of this Purification block. The module consists of 2 consecutive parts of feature extraction where each part includes one bottleneck layer and a 3×3 convolutional layer. The former is employed to reduce the number of channels from $D$ to $D/2$. The latter is employed to extract contextual features. Output of the second bottleneck layer is fed to another CORE module for refining localization information at the shallower scale.

### 3.3. BiFusion Feature Pyramid

As described before, the SoTA hourglass-based methods adopt a top-down path to generate a three-scale FP for object prediction by bringing semantic features from the deepest layer to other shallow layers. This top-down path cannot preserve the localization information precisely due to the quantization effect of pooling. To circulate semantic and localization information from a bottom-up pathway, current bidirectional methods adopt a memory- and-bandwidth consuming way to create new feature maps from shallow layers for feature fusion to predict object candidates more accurately. Different from these methods, this work reuses feature maps only from three layers (previous, current, and next) of the backbone by recursively performing the **CORE** and **Purification** modules. Figure 1(c) shows the recursive architecture to construct a bifusion feature pyramid. The output of the ($i$-1)th CORE and Purification modules is the input of the $i$th CORE module to generate more semantic contexts. The "re-using" mechanism makes it be memory- and-bandwidth efficient, and suitable for embedded applications. Recursively circulating semantic and localization information bi-directionally from deep and shallow layer also significantly improves the accuracy of small object detection. The recursive nature also brings localization information to refine the positions of large objects.

### 3.4. Residual Feature Pyramid

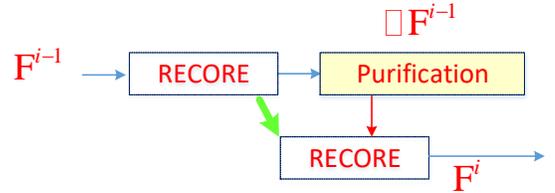

Figure 4. The residual version of the "CORE" module.

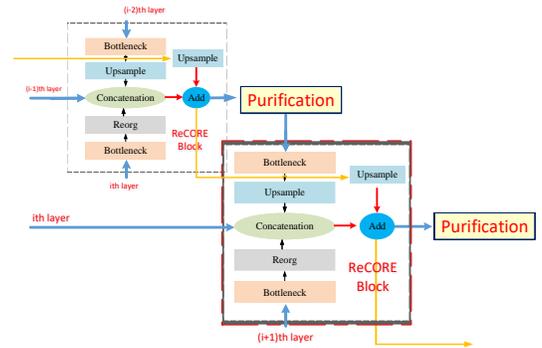

Figure 5. The "ReCORE" module.

Inspired by the concept of ResNet [5], the "residual" recursive formulation is adopted in this paper to implement and train the bifusion feature pyramid network. By recursively injecting the output of the ($i$-1)th CORE module to the $i$th CORE module, the architecture in Figure 1(c) can be converted to Figure 1(d) where the CORE module is changed to a residual version, *i.e.*, the RECORE (Residual CORE) module. The relation between the RECORE module and the Purification module can be thought of as a residual form:

$$F^i = F^{i-1} + \Delta F^{i-1}, \qquad (1)$$

where $F^i$ and $\Delta F^i$ denote the outputs of the $i$th RECORE and Purification modules, respectively. Although both sides of Eq.(1) are not really equal, Eq. (1) inspires us to construct the "residual" version of the CORE module as Figure 5. With the RECORE module, this paper constructs a new "residual" feature pyramid to circulate semantic and localization information bi-directionally from both deep and shallow layers. The residual nature makes the new feature pyramid be easily trained and integrated into different backbone, and thus significantly improves the results on small object detection.

### 3.5. Improvement on Large and Medium Object Detection

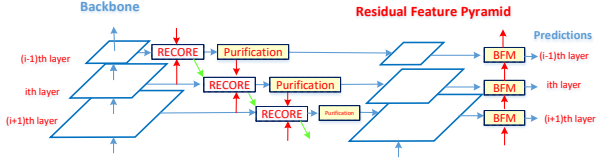

Figure 6. Overall architecture of the residual bi-fusion feature pyramid network.

As described before, YOLO V3 improved the accuracy of small object detection but resulted in inaccuracy on large sized object detection. To avoid this unexpected effects, the winner team in LPIRC 2019 [28] improved the results on large and medium sized objects but ignoring the ones on small objects to increase the overall accuracy. To achieve both high accuracies on small and large sized object detection, features among layers in the residual pyramid structure will further fused using a bottom-up pathway (shown in Figure 6).

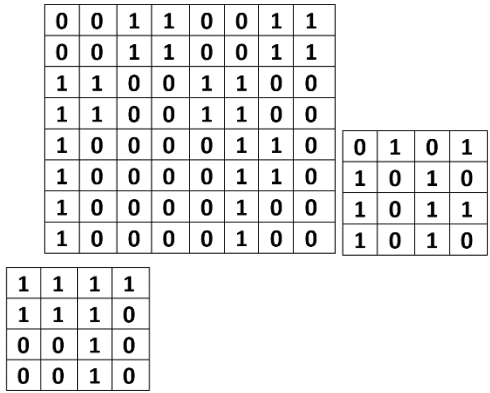

(a)
(b)    (c)
Figure 7. The pooling task will change the shift-invariance. Max-pooling on (a) results in (b). However, simply shifting the input in (a) and then max-pooling gives a different answer as (c).

Most of SoTA bidirectional methods use a concatenation operation to fuse features at the current layer and four sub-images, which are subsampled and reshaped from its shallow layer, to get a new feature map. However, the concatenation operation cannot reduce the effect of pooling to change shift invariance. In most recent works [19], [5], max-pooling has been the most commonly adopted method for executing down-sampling.

Consider the case shown in Figure 7. If one applies the max-pooling at the block shown in Figure 7(a), the result and block is shown in Figure 7(b) (kernel k=2, stride s=2). If one simply shifts Figure 7(a) with $\Delta x = 1$ and $\Delta y = 0$, a significantly different result will be produced ( as shown in Figure 7(c)). In this case, if one applied the max-pooling on the first two rows $\begin{bmatrix} 0,0,1,1,0,0,1,1 \\ 0,0,1,1,0,0,1,1 \end{bmatrix}$ of Figure 7(a) then the result will be [0,1,0,1] (the first row of in Figure 7(b)). However, if one position is shifted, the result of max-pooling will become [1,1,1,1] (see the first row in Figure 7(c)). It is obvious that a pooling operation definitely causes loss on shift-invariance requirement.

To reduce the effect of pooling, this paper suggests using a 1×1 convolution will be better than a concatenation operation to fuse features coming from both down-sampled data and those from the current layer. The 1×1 convolution filter used in the bi-fusion process is automatically learned; while the concatenation is "hand-crafted". Figure 8 shows the architecture of our bi-fusion module. Let $FM_i$ denote the $i$th feature map in the residual feature pyramid. From $FM_i$, we use a re-shaping technique to divide it into four sub-patches. Then, 1×1 convolution is applied to fuse four sub-patches and $FM_{i-1}$ to generate $C_{i-1}$ channels of features. In this example, the 1-D signal [0,0,1,1,0,0,1,1] will be split into {[0,1,0,1], [0,1,0,1]}. If the 1×1 convolution is "blurring", the result will be similar to the blur-pooling technique. For example, if $FM_{i-1}$ is [1,1,1,1] and the 1×1 convolution is "average", the fusion result of {[0,1,0,1], [0,1,0,1]} and {[1,1,1,1]} will be [1/3,1,1/3,1], which is close to [0,1,0,1] for the case that $FM_{i-1}$ = [0,1,0,1]. If the 1×1 convolution is "median" or "min", the fused features for [1,1,1,1] and [0,1,0,1] are the same. Extensive experiments show the 1×1 convolution performs much better than the concatenation operation on both small and large sized object detection. Figure 6 shows the overall architecture of the residual bi-fusion feature pyramid network. The BFM enhances the results on large-sized object detection more than small-sized object detection.

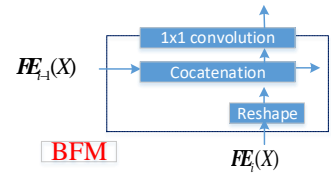

Figure 8. BF module for feature fusion on the residual feature pyramid.

### 4. Experimental Results

Model evaluations are conducted on the bounding box detection task of the MS COCO [40] benchmark using a machine with NVIDIA Titan X. ReBiF net is compared with the latest state-of-the-art object detectors in terms of accuracy and efficiency. Metric adopted for performance evaluation is Average Precision (AP). Inference time is represented as FPS (Frames per Second).

### 4.1. Accuracy Improvements by BFM

**Table 1. Ablation study of BFM across different backbones.**

| Backbone | BFM | FPS | AP | $AP_{50}$ | $AP_{75}$ | $AP_S$ | $AP_M$ | $AP_L$ |
|---|---|---|---|---|---|---|---|---|
| DarkNet 53 |  | 28.9 | 28.6 | 50.7 | 29.6 | 15.5 | 30.4 | 35.3 |
| 515x512 | ✓ | 28.4 | 34.9 | 57.2 | 37.7 | 18.6 | 37.1 | 45.3 |
| Pelee |  | 85.8 | 26.7 | 49.9 | 26.2 | 13.5 | 27.8 | 33.5 |
| 512x512 | ✓ | 84.5 | 28.3 | 51.8 | 28.4 | 14.0 | 30.1 | 35.6 |
| VGG 16 |  | 42.4 | 34.1 | 58.3 | 35.8 | 17.9 | 35.9 | 44.1 |
| 512x512 | ✓ | 42 | 34.6 | 58.6 | 36.7 | 18.6 | 36.5 | 44.3 |
| DenseNet201 |  | 40.2 | 31.3 | 54.5 | 32.5 | 15.7 | 33.8 | 40.8 |
| 512x512 | ✓ | 39.4 | 31.5 | 54.7 | 33.3 | 15.9 | 33.9 | 41.1 |

Since ReBiF net is developed specifically for both small and large sized object detection, we evaluate the effects of BFM on object detection based on the MS COCO dataset across different backbones. Table 1 tabulates the ablation studies to show the advantage of BFM. Four backbones were compared to demonstrate the generalization of BFM; that is, PeLee [45], DarkNet-53 [2], VGG16 [19], and DenseNet [17]. Table 1 tells us the computation load of BFM is light and can be ignored for all backbones. We also observe its generalization to improve the accuracy of object detection under different backbones across different object sizes. An important phenomenon found from Table 1 is: the accuracy of a light backbone is improved with a larger margin than a deep one. The improvements on AP50 with BFM for DarkNet, Pelee, and DenseNet are 6.5%, 3.5%, and 0.2%, respectively. It indicates that BFM improves the results on large objects more than small objects. BFM will be a good solution to improve the accuracy loss of large sized object detection if more attentions are focused on small object objection.

### 4.2. Improvements by ReBiF net (BFM+ReCORE)

**Table 2. Ablation study of ReCORE and BFM**

| Backbones | ReCORE | BFM | FPS | AP | $AP_{50}$ | $AP_{75}$ | $AP_S$ | $AP_M$ | $AP_L$ |
|---|---|---|---|---|---|---|---|---|---|
| Pelee* |  |  | 110 | 23.6 | 45.3 | 22.5 | 7 | 25.6 | 34.9 |
| Pelee | ✓ |  | 106 | 23.9 | 46 | 22.5 | 8.1 | 24.4 | 34.7 |
| 416x416 | ✓ | ✓ | 104 | 26.7 | 49.5 | 26.3 | 10.3 | 28 | 37.4 |
| Darknet53** |  |  | - | 32 | 56.5 | 33 | 17.4 | 34 | 41.4 |
| Darknet53*** |  |  | - | 28.6 | 50.7 | 29.6 | 15.5 | 30.4 | 35.3 |
| Darknet53 | ✓ |  | 44 | 36 | 59.5 | 38.2 | 18.9 | 37.3 | 47.1 |
| Darknet53 |  | ✓ | 44 | 34.9 | 57.2 | 37.7 | 18.6 | 37.1 | 45.3 |
| 512x512 | ✓ | ✓ | 42 | 36.8 | 59.7 | 39.6 | 19 | 39.5 | 48 |

\* Trained and tested by ourselves according to the paper.
\*\* Test results with weights provided in the official website of YOLOv3.
\*\*\* Trained and tested by ourselves according to the instruction.

Another novelty of this work is the design of ReCORE module. With BFM and the ReCORE module, we proposed the ReBiF net for object detection. When Darknet 53 was adopted as the backbone, YOLO v3 will be a good candidate for performance comparisons. Table 2 shows the ablation study of ReBiF net and YOLO V3. From the table, we can see that ReBiF net outperforms all categories. It is noted that the frame rate difference between before/after using the BFM is minor. Moreover, when the input size is 608×608, YOLOv3 [2] with BFM outperforms YOLOv3 on all categories. As to the input size with 512×512, YOLOv3 with BFM outperforms YOLOv3 without BFM on all categories. The highest improvements are made on the small sized object detection. For small objects, BFM brings an increasing trend over the input size progression. On the contrary, improvements on the large objects have decreasing trend over the input size progression.

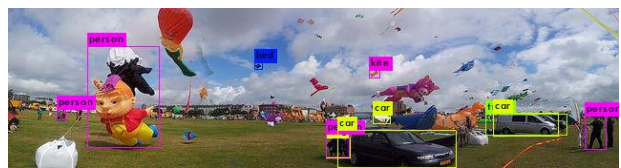
(a) YOLOv3_512x512

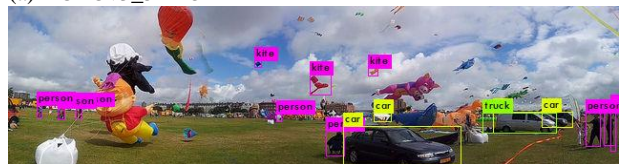
(b) YOLOv3+BFM_512x512

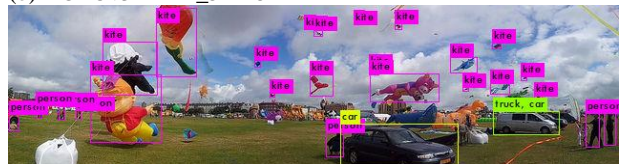
(c) ReBiF512x512

Figure 9. Small Object detection results on image 000000343548.jpg of MS COCO test_dev set.

Figure 9 shows the comparisons of object detection among three models on an image (with input size 512×512) selected from MS COCO test-dev set. These models are a) YOLOv3, b) YOLOv3+BFM, c) ReBiF net with BFM. It is obvious that the last one receives the best detection results.

### 4.3. Improvements by Residual Feature Pyramid Network

To compare the differences between our residual FPN and the original FPN, we can check the effect of FPN on accuracy improvement when the number of feature layers increases. Three layers are commonly adopted in most FPN architectures due to little accuracy improvement when more than three layers are used. However, our residual FPN can still improve the accuracy of object detection significantly when four or five layers are used. To prove it,
we employed VOC 2012 trainval set for training the ReFPN and general FPN with 50K iteration as described in [3]. Validations are made on an input size of 416×416. Training is performed on GPU Nvidia Titan X, and testing is on Nvidia Jetson TX2. Table 3 tabulates the performance comparisons when feature layers increase. Clearly, the accuracy of the FPN decreases when four layers are used. However, for the residual FPN, its accuracy still increases. When five layers are used, the residual FPN performs the best, with mAP value 74.9%, which is higher than Pelee (70.9%). In addition, even though more layers are used, its fps is better than Pelee which is a light backbone for mobile applications.

**Table 3. Ablation study of Residual Feature Pyramid on PASCAL VOC 2012 test set.**

| Number of Scale | | | | FPN | ReFPN | FPS | Bflops | Model Size | mAP |
|---|---|---|---|---|---|---|---|---|---|
| 2 | 3 | 4 | 5 | | | | | | |
| ✓ | | | | ✓ | | 33.73 | 3.992B | 12.4M | 69.65 |
| ✓ | | | | | ✓ | 33.57 | 3.992B | 12.4M | 70.10 |
| | ✓ | | | ✓ | | 32.41 | 4.228B | 12.6M | 70.08 |
| | ✓ | | | | ✓ | 31.02 | 4.254B | 12.6M | 70.70 |
| | | ✓ | | ✓ | | 28.42 | 4.490B | 12.6M | 68.41 |
| | | ✓ | | | ✓ | 26.45 | 4.683B | 12.7M | 70.92 |
| | | | ✓ | ✓ | | 24.04 | 4.982B | 13.0M | 69.12 |
| | | | ✓ | | ✓ | 22.32 | 5.231B | 13.2M | 74.9 |
| Pelee | | | | | | 14.71 | 2.420B | 5.98M | 70.90 |

4.4. Learning Rich information with lightweight backbone.

**Table 4. Ablation Study with SoTA lightweight backbones on Jetson TX2.**

| Method | Backbone | Input size | FPS* | AP | $AP_{50}$ | $AP_{75}$ | $AP_S$ | $AP_M$ | $AP_L$ |
|---|---|---|---|---|---|---|---|---|---|
| Yolo v3-tiny | tiny | 416 | 37 | - | 33.1 | - | - | - | - |
| Pelee | Pelee | 304 | 14 | 22.4 | 38.3 | 22.9 | - | - | - |
| PRN[46] | Pelee | 416 | 27 | - | 45 | - | - | - | - |
| ReCORE | Pelee | 416 | 24 | 23.9 | 46 | 22.5 | 8.1 | 24.4 | 34.7 |
| ReBiF | Pelee | 416 | 23 | **26.7** | **49.5** | **26.3** | **10.3** | **28** | **37.4** |

*FPS including preprocessing, model inference, and post-processing time.

Our ReBiF architecture is suitable for lightweight backbones. Three lightweight backbones were compared; that is, YoloV3-tiny, PeLee and PRN[46]. Table 4 tabulates their performances on Nvidia Jetson TX2. The Pelee backbone is chosen as the baseline to prove the superiority of the ReBiF on embedded applications. ReBiF outperforms all SoTA lightweith backbones.

4.5. Comparisons with state-of-the-art models

Table 4 tabulates the comparisons with SoTA models. To compare the efficiency and accuracy of our ReBiF net with state-of-the-art models, the inference time is calculated for a single image by taking the sum of the CNN time and NMS (non-maximum suppression) time of 999 random images, and divide by 999. Inference time vs. APs curve is shown in Figure 10. Moreover, we only compare the state-of-the-art models which have an inference time <100ms (≥10FPS) for real-time applications. ReBiF net has the advantage of bidirectional fusion of multi-scale contextual features and computationally low yet feature preserving operations; therefore, it achieves outstanding speed-accuracy curve compared with state-of-the-arts. When both accuracy and speed are considered, ReBiF outperforms the latest state-of-the-art one-stage object detectors, e.g., LRF [7], M2Det, and RetinaNet. Although M2Det [25] outperfored other methods on all multi-scale categories on MS COCO, its model is complicated and time-consuming for a real-time object detection task.

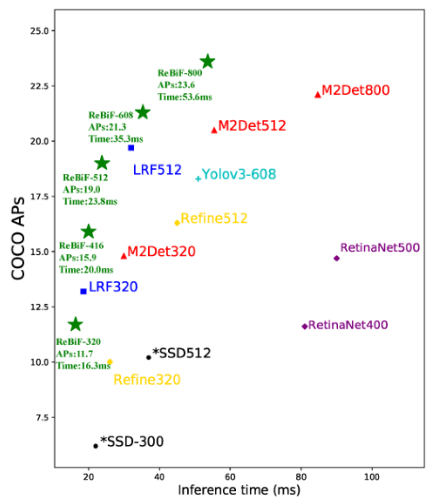

Figure 10. $AP_S$ vs. Inference Time Curve. ReBiF nets are tested on NVIDIA Titan X.

5. Discussions and Conclusions

A new structure "residual feature pyramid" was proposed in this paper. It is bidirectional to fuse both deep and shallow features towards more effective and robust detection for both small sized and large sized objects. Due to the "residual" nature, it can be easily trained and integrated to different backbones (even deeper or lighter) than other bi-directional methods. From the ablation study, accuracy improvement is still found even more than three layers are adopted in this pyramid. To avoid the problem of handcrafted anchors, some anchor-free method will be develiped to further imporve the detection accruacy.

Table 5. Comparisons on MS COCO test-dev set with SoTA models

| Method | Backbone | Input size | FPS | AP | $AP_{50}$ | $AP_{75}$ | $AP_S$ | $AP_M$ | $AP_L$ |
|---|---|---|---|---|---|---|---|---|---|
| Faster R-CNN [30] | VGGNet-16 | ~1000x600 | 7 | 21.9 | 42.7 | - | - | - | - |
| R-FCN [19] | ResNet-101 | ~1000x600 | 9 | 29.9 | 51.9 | - | 10.8 | 32.8 | 45 |
| Faster R-CNN w/ FPN [47] | ResNet-101-FPN | ~1000x600 | 6 | 36.2 | 59.1 | 39 | 18.2 | 39 | 48.2 |
| Cascade R-CNN [44] | ResNet-101-FPN | ~1280x800 | 7 | 42.8 | 62.1 | 46.3 | 23.7 | 45.5 | 55.2 |
| Mask-RCNN [42] | ResNet-101-FPN | ~1280x800 | 5 | 39.8 | 62.3 | 43.4 | 22.1 | 43.2 | 51.2 |
| SNIP [53] | DPN-98 | - | - | 45.7 | 67.3 | 51.1 | 29.3 | 48.8 | 57.1 |
| Deformable R-FCN [31] | ResNet-101 | ~1000x600 | 8 | 34.5 | 55 | - | 14 | 37.7 | 50.3 |
| SSD-300 [3] | VGGNet-16 | 300x300 | 43 | 25.1 | 43.1 | 25.8 | 6.6 | 25.9 | 41.4 |
| SSD [3] | ResNet-101 | 321x321 | 50 | 28 | 45.4 | 29.3 | 6.2 | 28.3 | 49.3 |
| DSSD [51] | ResNet-101 | 321x321 | - | 28 | 46.1 | 29.2 | 7.4 | 28.1 | 47.6 |
| CFENet-300 [50] | VGGNet-16 | - | - | 30.2 | 50.5 | 31.3 | 12.7 | 32.7 | 46.6 |
| RefineDet-320 [15] | VGGNet-16 | 320x320 | 38.7 | 29.4 | 49.2 | 31.3 | 10 | 32 | 44.4 |
| RefineDet-320 [15] | ResNet-101 | 320x320 | - | 32 | 51.4 | 34.2 | 10.5 | 34.7 | 50.4 |
| RFBNet [52] | VGGNet-16 | 300x300 | 67 | 30.3 | 49.3 | 31.8 | 11.8 | 31.9 | 45.9 |
| EFIP[55] | VGGNet-16 | 300x300 | 71 | 30 | 48.8 | 31.7 | 10.9 | 32.8 | 46.3 |
| PFPNet-R320 [9] | VGGNet-16 | 320x320 | 33 | 31.8 | 52.9 | 33.6 | 12 | 35.5 | 46.1 |
| M2Det-320 [25] | VGGNet-16 | 320x320 | 33.4 | 33.5 | 52.4 | 35.6 | 14.4 | 37.6 | 47.6 |
| M2Det-320 [25] | ResNet-101 | 320x320 | 21.7 | 34.3 | 53.5 | 36.5 | **14.8** | 38.8 | 47.9 |
| LRFNet [49] | VGGNet-16 | 300x300 | **77** | 32 | 51.5 | 33.8 | 12.6 | 34.9 | 47 |
| LRFNet [49] | ResNet101 | 300x300 | 53 | 34.3 | 54.1 | 36.6 | 13.2 | 38.2 | **50.7** |
| **ReBiF [Ours]** | VGGNet-16 | 320x320 | 54 | 30.4 | 53 | 31.3 | 11.5 | 31.7 | 45.3 |
| **ReBiF-320 [Ours]** | DarkNet-53 | 320x320 | 61 | 32.1 | 53.4 | 33.8 | 11.7 | 33.6 | 48.9 |
| **ReBiF-320 [Ours]** | ResNet101 | 320x320 | 42.8 | **34.5** | **55.2** | **38.0** | 14.0 | **38.9** | 50.5 |
| YOLOv2 [4] | DarkNet-19 | 544x544 | 40 | 21.6 | 44 | 19.2 | 5 | 22.4 | 35.5 |
| YOLOv3-608 [2] | DarkNet-53 | 608x608 | 19.8 | 33 | 57.9 | 34.4 | 18.3 | 35.4 | 41.9 |
| SSD-512 [3] | VGGNet-16 | 512x512 | 22 | 28.8 | 48.5 | 30.3 | 10.9 | 31.8 | 43.5 |
| SSD-512 [3] | ResNet101 | 513x513 | 31.3 | 31.2 | 50.4 | 33.3 | 10.2 | 34.5 | 49.8 |
| DSSD [51] | ResNet101 | 513x513 | 6.4 | 33.2 | 53.3 | 35.2 | 13 | 35.4 | 51.1 |
| RefineDet-512 [15] | VGGNet-16 | 512x512 | 22.3 | 33 | 54.5 | 35.5 | 16.3 | 36.3 | 44.3 |
| RefineDet-512 [15] | ResNet101 | 512x512 | - | 36.4 | 57.5 | 39.5 | 16.6 | 39.9 | 51.4 |
| Rev-Dense [54] | VGGNet-16 | 512x512 | - | 31.2 | 52.9 | 32.4 | 15.5 | 32.9 | 43.9 |
| CFENet-512 [50] | VGGNet-16 | - | - | 34.8 | 56.3 | 36.7 | 18.5 | 38.4 | 47.4 |
| RFBNet [52] | VGGNet-16 | 512x512 | 33 | 33.8 | 54.2 | 35.9 | 16.2 | 37.1 | 47.4 |
| RFBNet-E [52] | VGGNet-16 | 512x512 | 30 | 34.4 | 55.7 | 36.4 | 17.6 | 37 | 47.6 |
| RetinaNet [13] | ResNet-101-FPN | ~832x500 | 11 | 34.4 | 55.7 | 36.8 | 14.7 | 37.1 | 47.4 |
| RetinaNet+AP-Loss [13] | ResNet-101-FPN | 512x512 | 11 | 37.4 | 58.6 | 40.5 | 17.3 | 40.8 | 51.9 |
| EFIP [55] | VGGNet-16 | 512x512 | 34 | 34.6 | 55.8 | 36.8 | 18.3 | 38.2 | 47.1 |
| PFPNet-S512 [9] | VGGNet-16 | 512x512 | 24 | 33.4 | 54.8 | 35.8 | 16.3 | 36.7 | 46.7 |
| PFPNet-R512 [9] | VGGNet-16 | 512x512 | 24 | 35.2 | 57.6 | 37.9 | 18.7 | 38.6 | 45.9 |
| M2Det-512 [25] | VGGNet-16 | 512x512 | 18 | 37.6 | 56.6 | 40.5 | 18.4 | 43.4 | 51.2 |
| M2Det-512 [25] | ResNet-101 | 512x512 | 15.8 | **38.8** | 59.4 | **41.7** | **20.5** | 43.9 | **53.4** |
| LRFNet [49] | VGGNet-16 | 512x512 | 38 | 36.2 | 56.6 | 38.7 | 19 | 39.9 | 48.8 |
| LRFNet [49] | ResNet-101 | 512x512 | 31 | 37.3 | 58.5 | 39.7 | 19.7 | 42.8 | 50.1 |
| **ReBiF [Ours]** | VGGNet-16 | 512x512 | 28.9 | 36.8 | 59.7 | 39.5 | 19.1 | 38.9 | 48.7 |
| **ReBiF-512 [Ours]** | DarkNet-53 | 512x512 | **42** | 36.8 | 59.7 | 39.6 | 19 | 40.5 | 48 |
| **ReBiF-512 [Ours]** | ResNet-101 | 512x512 | 28.4 | 38 | **60.3** | 40.3 | 20.1 | 43.1 | 52.1 |